\documentclass[letterpaper]{article} 
\usepackage{aaai25}  
\usepackage{times}  
\usepackage{helvet}  
\usepackage{courier}  
\usepackage[hyphens]{url}  
\usepackage{graphicx} 
\usepackage{natbib}  
\usepackage{caption} 
\usepackage{algorithm}
\usepackage{algorithmic}

\usepackage{amsmath}
\usepackage{amssymb}
\usepackage{booktabs}
\usepackage{array}
\usepackage{multirow}
\usepackage{pifont}
\usepackage{cite}
\usepackage{mathabx}
\usepackage{colortbl}
\definecolor{lightgray}{gray}{0.9}
\usepackage{xcolor}

\usepackage{newfloat}
\usepackage{listings}
\DeclareCaptionStyle{ruled}{labelfont=normalfont,labelsep=colon,strut=off} 
\floatstyle{ruled}
\newfloat{listing}{tb}{lst}{}
\floatname{listing}{Listing}
\title{WiFi CSI Based Temporal Activity Detection via Dual Pyramid Network}
\author{
    Zhendong Liu\textsuperscript{\rm 1},
    Le Zhang\textsuperscript{\rm 1}~\thanks{Corresponding author},
    Bing Li\textsuperscript{\rm 1},
    Yingjie Zhou\textsuperscript{\rm 2},
    Zhenghua Chen\textsuperscript{\rm 3},
    Ce Zhu\textsuperscript{\rm 1}
}

\affiliations{

    \textsuperscript{\rm 1}School of Information and Communication Engineering, University of Electronic Science and Technology of China\\
    \textsuperscript{\rm 2}College of Computer Science, Sichuan University\\
    \textsuperscript{\rm 3}Institude for Infocomm Research, Agency for Science, Technology and Research (ASTAR), Singapore\\
    
    lzdjohn@std.uestc.edu.com, \{lezhang, bing\_li, eczhu\}@uestc.edu.cn, yjzhou@scu.edu.cn, chen\_zhenghua@i2r.a-star.edu.sg
}

\begin{document}
\maketitle
\begin{abstract}
We address the challenge of WiFi-based temporal activity detection and  propose an efficient Dual Pyramid Network that integrates Temporal Signal Semantic Encoders and Local Sensitive Response Encoders. The Temporal Signal Semantic Encoder splits feature learning into high and low-frequency components, using a novel Signed Mask-Attention mechanism to emphasize important areas and downplay unimportant ones, with the features fused using ContraNorm. The Local Sensitive Response Encoder captures fluctuations without learning. These feature pyramids are then combined using a new cross-attention fusion mechanism. We also introduce a dataset with over 2,114 activity segments across 553 WiFi CSI samples, each lasting around 85 seconds. Extensive experiments show our method outperforms challenging baselines.
\end{abstract}

\begin{links}
    \link{Code}{https://github.com/AVC2-UESTC/WiFiTAD}
\end{links}

\section{Introduction}
\label{sec:intro}
Using IoT devices to recognize human activities like walking, falling, and lying down has numerous applications~\cite{kong2022human}. Recently, there has been growing interest in not just short-term activity analysis but also long-term daily behavior monitoring, which is vital for real-world applications like health monitoring and medical statistics~\cite{8761602}.

For long-term monitoring, researchers are increasingly focusing on processing extended data streams, specifically through Temporal Activity Detection (TAD), which aims to automatically identify activities and their timing within prolonged monitoring data~\cite{shou2016temporal}. However, most of these efforts rely on camera sensors, which require a direct line of sight (LOS) to function effectively. This limitation restricts their use in low-light conditions and raises significant privacy concerns, making them less suitable for sensitive environments where confidentiality is important~\cite{sun2022human}.

As a privacy-preserving alternative, researchers have turned to non-invasive technologies like WiFi Channel State Information (CSI) to sense and recognize human activities~\cite{chen2018wifi}. Unlike cameras, WiFi CSI doesn’t capture visual images, eliminating privacy issues. By detecting the multi-path effects caused by human movements, CSI allows for Non-Line-of-Sight (NLOS) monitoring along fixed transmitter-receiver paths. The widespread availability of WiFi makes this approach not only cost-effective but also highly accurate for activity recognition, even in challenging conditions~\cite{10.1145/3264958}.

\begin{figure}[!t]
    \centering
    \vspace{-0.3cm}
    \includegraphics[width=0.9\linewidth]{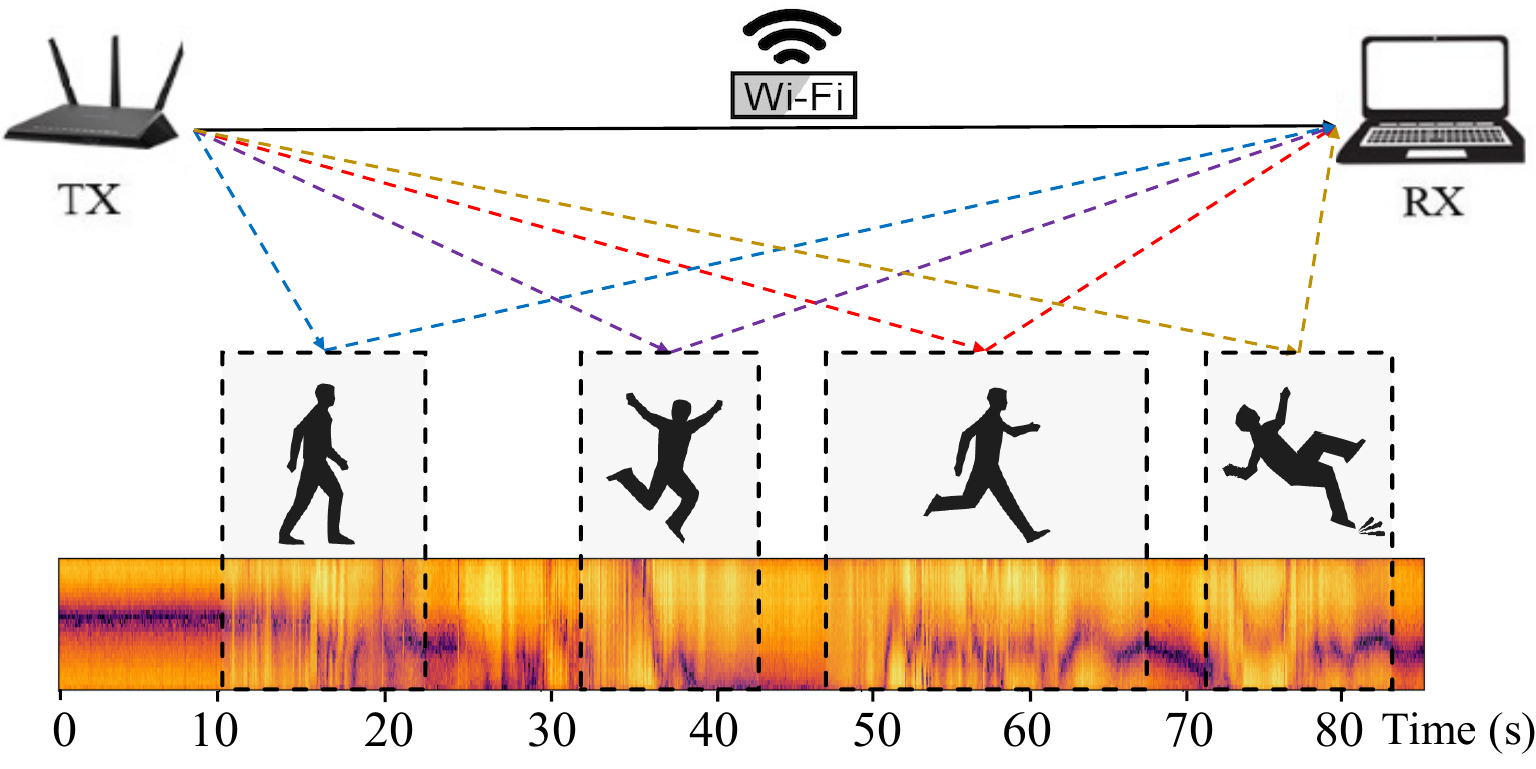}
    \caption{The variation of ubiquitous wireless signal caused by different human activities along the temporal axis.}
    \vspace{-0.3cm}
    \label{fig:wireless}
\end{figure}

Despite significant progress in using CSI for human activity recognition (HAR), most existing methods assume that input signals are pre-segmented into distinct activities, focusing on correctly identifying activities from a predefined set of classes~\cite{chen2018wifi,meng2023graphar,li2021two,chavarriaga2013opportunity,xiao2020deepseg,yousefi2017survey}. While effective for well-defined segments, these methods struggle in more complex scenarios where activity boundaries are not predefined, and the signal is continuous and untrimmed, as shown in~Fig.~\ref{fig:wireless}.

Temporal activity detection is well-developed in computer vision using visual inputs~\cite{wang2023temporal}, but applying these methods directly to WiFi CSI-based activity detection is challenging due to differences in data modality and characteristics. Unlike visual data, which provides rich spatial features and clear temporal sequences, WiFi CSI data is primarily temporal, noisy, and lacks intuitive spatial cues. Environmental factors introduce noise in WiFi signals, which is different from the noise typically encountered in visual data. Additionally, WiFi CSI suffers from a scarcity of annotated data, unlike the large labeled datasets available in computer vision. Moreover, WiFi CSI data, generated by inexpensive sensors, requires more efficient models than the computationally intensive approaches used in computer vision. These differences highlight the need for tailored methods for WiFi CSI, making it unsuitable to directly apply computer vision-based approaches.

To address these challenges, we explore wireless Temporal Activity Detection and introduce DPWiT, an end-to-end learning model. DPWiT uses a multi-scale dual pyramid structure that combines frequency-aware feature learning with fluctuation information to accurately identify activities and their precise locations within untrimmed, long-term signals. The core component, the Dual Pyramid Temporal Context Modeling (DPTCM), generates multi-scale features through Temporal Signal Semantic Encoders (TSSE) and Local Sensitive Response Encoders (LSRE), which are fused using Cross-attention Pyramid Fusion modules. We also collected and annotated a comprehensive untrimmed WiFi CSI dataset covering seven daily activities: walk, run, jump, wave, fall, sit, and stand. This dataset includes 553 untrimmed samples with 2,114 activity instances, each annotated with start time, end time, and category. To summarize, we contribute in:

\begin{itemize}
\item We systematically study the WiFi based Temporal Activity Detection (TAD) task and introduce a comprehensive solution. This includes developing a novel method, creating a real-world dataset of long-term, untrimmed multi-activity wireless signals, and establishing a new benchmark for future research.
\item We explore the classification and localization sub-tasks of TAD, finding that high-frequency information is crucial for localization, while low-frequency information is better for identifying activity categories.
\item We propose DPWiT, a model that combines frequency-aware learning with dual pyramid fusion, achieving state-of-the-art results on real-world datasets. For low-frequency learning, we introduce Signed Mask-Attention, which better highlights important areas and downplays unimportant ones, enhancing the model's focus on critical regions.
\end{itemize}

\section{Related Work}
\label{sec:related}

\subsection{CSI based Huam Activity Analysis}

Recently, deep learning-based strategies have been increasingly applied to CSI-based human activity recognition, inspired by the success of deep neural networks in various fields~\cite{krizhevsky2012imagenet, li2021two}. Models like ABLSTM~\cite{chen2018wifi} and Transformers~\cite{li2021csi} have demonstrated the advantages of temporal context modeling in improving recognition accuracy. Some studies have extended WiFi signal use to long-term monitoring, achieving effective respiration monitoring in real home environments~\cite{tian2018rf, 10.1145/3448092, wang2023u}. Although they analyze successive activity recognition from the temporal angle, these methods are generally limited to single, well-defined activity segments and regular patterns. The challenge of accurately detecting and recognizing various activities from untrimmed, unrestricted WiFi CSI data remains largely unresolved.

\begin{figure*}[!t]
    \centering
    \includegraphics[scale=0.89]{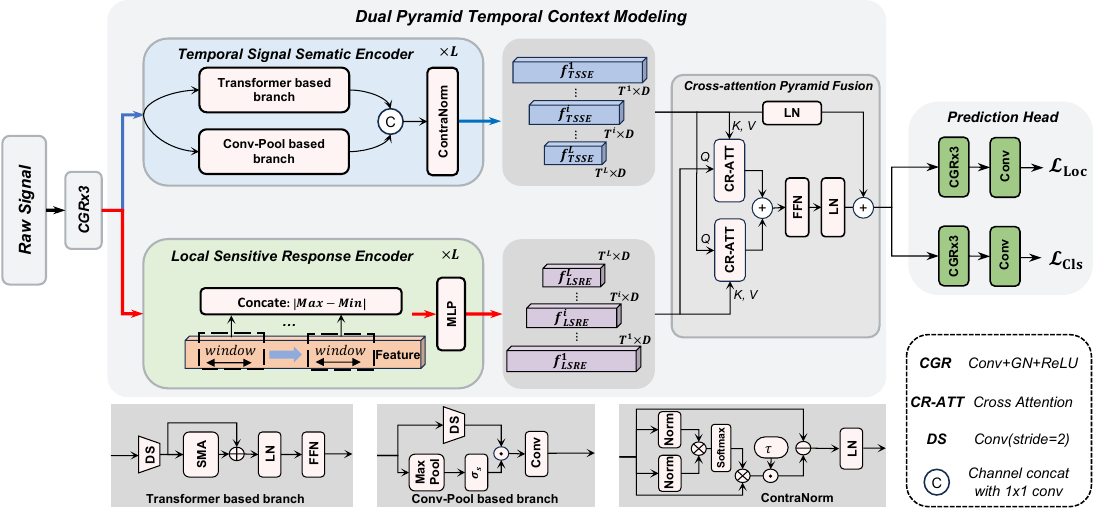}
    \caption{Overview of our network. Given a raw signal input, we employ 3 CGR layers to project the signal and generate output feature through dual pyramid temporal context modeling. Finally, the features are processed by the prediction head to transform into TAD results.}
    \vspace{-0.5cm}
    \label{fig:framework}
\end{figure*}

\subsection{Temporal Activity Detection}
Temporal activity detection is well-developed in computer vision using visual inputs~\cite{wang2023temporal}. Existing vision-based TAD methods are divided into one-stage and two-stage approaches. Two-stage methods~\cite{chen2022dcan, xia2022learning} first generate potential instance proposals and then classify and refine them using independently trained detectors. In contrast, one-stage methods~\cite{shi2023tridet, yang2020revisiting} use an end-to-end pipeline to simultaneously localize and recognize actions. Although temporal action detection (TAD) is well-studied in the vision community, WiFi-based TAD remains in its early stages. It requires precise modeling of signal temporal boundaries and an in-depth understanding of multiple actions. However, as demonstrated in our experiments, directly applying vision-based methods to WiFi CSI signals often yields suboptimal results due to the unique characteristics of wireless signals.

\section{Method}
\label{sec:method}
\subsection{Problem Definition}
Given a dataset of long-term signals $\mathcal{D} = \{X_i\}_{i=1}^{n}$, where each signal instance $X_i$ contains $M_i$ action segments $Y_i = \{(s_m, e_m, c_m)\}_{m=1}^{M_i}$, with $s_m$ representing the start time, $e_m$ the end time, and $c_m$ the corresponding action category, our task is to detect all action segments in $Y_i$ based on the input signal $X_i$.

\subsection{Some Preliminary Results}
Before delving into the detailed design of our proposed methods, we present some preliminary results that motivate our solution.

We empirically observed that CSI signals are highly complex, as illustrated in Fig.~\ref{fig:visual}, primarily due to the inherently noisy characteristics caused by multi-path effects~\cite{yang2013rssi}. Despite this complexity, our task involves identifying the precise start and end times of each potential activity. We hypothesize that these temporal boundaries are largely driven by rapid changes in the signal, which can be effectively captured by high-frequency information. Additionally, identifying the specific activity category requires a comprehensive understanding of the semantic information across an entire input segment, which can be modeled by low-frequency information. Low-frequency components are more effective for distinguishing between different types of activities, as they capture the broader, more stable patterns in the signal that are crucial for accurate classification within the detected temporal boundaries. Similar approaches to signal analysis from a frequency perspective has also been validated in the vision community~\cite{wang2022antioversmoothing,li2023feature}.

Existing work~\cite{wang2022antioversmoothing, li2023feature} has shed light on network design from a frequency perspective. Self-attention, by focusing on all parts of the input sequence, tends to aggregate information across the entire sequence, smoothing out variations and emphasizing the global structure. This behavior is akin to a low-pass filter in signal processing, making self-attention particularly effective at extracting low-frequency features that capture broader, more stable patterns—ideal for tasks requiring a global context or semantic understanding. Convolutional layers, on the other hand, especially with small kernel sizes, are designed to capture local patterns by applying filters over small receptive fields. These operations are sensitive to sharp transitions and fine details, characteristic of high-frequency components. When combined with pooling, which down-samples the input, convolutional layers further amplify high-frequency features, making them effective at detecting edges, textures, and other fine details. 

To validate our hypothesis, we conducted preliminary studies from a frequency perspective. In our first case study, we designed two networks: one featuring a transformer backbone and the other a convolutional network backbone. Both networks were configured with comparable numbers of parameters and employed the same classification and localization heads. We assessed these networks using two metrics: localization mean Intersection over Union (mIoU), classification precision. The mIoU measures the average IoU between the predicted and ground truth (GT) boundaries, regardless of the activity labels, evaluating the network's capability to detect potential activities within the input signal. Precision is defined as the ratio of correctly predicted activity labels—those whose predicted boundaries overlap with the GT boundaries within a predefined tIoU range of [0.3:0.7:0.1]—to the total number of predictions. Precision evaluate the network's ability to accurately understand the context of the input signal, assuming that rough segmentation has already been achieved. The results, summarized in Table~\ref{tab:freq}, indicate that the transformer-based solution, which focuses on learning global dependencies and capturing low-frequency semantic information, excels in classifying activity categories. Conversely, the convolutional network-based solution, adept at capturing local patterns and high-frequency features, performs better in identifying activity boundaries.

To further validate this phenomenon, we conducted additional experiments. Initially, we transformed the input signal into the frequency domain using the Fast Fourier Transform (FFT). We then identified the cutoff frequency at the point where the power spectrum decreased by 6 dB and divided the frequency spectrum into two parts: low-frequency and high-frequency, based on this cutoff point. Subsequently, we obtained the low-frequency and high-frequency signals by transforming the respective frequency bands back to the time domain using the Inverse FFT (IFFT). We then input these transformed signals into our proposed model, details of which will be elaborated in the following section, and evaluated their results. The outcomes are consistent with our previous study, further confirming the role of different frequency components in the task of Temporal Activity Detection. More results could be found in the supplementary material in our code.




 \begin{table}[t]
\centering
\footnotesize
\begin{tabular}{c|c|c}
\toprule
\textbf{Network} & \textbf{mIoU} & \textbf{Precision}  \\ 
\midrule
Transformer  &  47.1 & 21.9 \\
Convolutional Network & 49.4 & 18.3  \\ 
\midrule
\textbf{Frequency Band} & \textbf{mIoU} & \textbf{Precision} \\ 
\midrule

Low-Frequency Inputs & 43.5 & 19.1   \\ 
High-Frequency Inputs & 44.6 & 15.8 \\
\bottomrule
\end{tabular}
\caption{Preliminary Results and Analysis from the frequency perspective.}
\vspace{-0.5cm}
\label{tab:freq}
\end{table}

\subsection{Model Overview}

Given the distinct roles of frequency components in Temporal Activity Detection (TAD), we propose a frequency-aware learning framework. As shown in Figure \ref{fig:framework}, the input signal is first processed through three CGR (Conv+GroupNorm+ReLU) layers. The resulting features are then passed to a dual pyramid temporal context modeling module, which includes $L\times$ Temporal Signal Semantic Encoders (TSSE) and Local Sensitive Response Encoders (LSRE), followed by a cross-attention pyramid fusion module. The TSSE consists of a transformer branch and a Conv-Pool branch, designed to learn low and high frequencies, respectively. These features are integrated via a ContraNorm module. The LSRE captures signal fluctuations in a learning-free manner, and the features from both encoders are aligned through a cross-attention pyramid fusion mechanism. Finally, a prediction head outputs the detection results for training and inference.

\subsection{Dual Pyramid Temporal Context Modeling}
\label{sec:backbone}
The feature encoders starts from the projected feature $f\in \mathbb{R}^{T \times D}$, where $T$ represents the signal timestamp points and $D$ represents the channels. Through the local sensitive response encoder and temporal signal semantic encoders, two multi-scale feature pyramids are created.

\subsubsection{Temporal Signal Semantic Encoder}

We designed two distinct network branches, each featuring core modules of self-attention and convolutional-pooling operations, tailored to preferentially learn low and high frequencies, respectively. 

More specifically, given a feature $f \in \mathbb{R}^{T \times D}$, we employ two branches to process the feature. The first branch is transformer-based, where we introduce a novel Signed Mask-Attention (SMA) mechanism to enhance the extraction of low-frequency features in the signal. These low-frequency features serve as crucial cues for achieving a comprehensive understanding of the semantic information across the entire segment of the input. For the input feature $f$, we divide it into multi-heads. For the \textit{i}-th head $f_\textit{i} \in \mathbb{R}^{T \times d_k}$, we compute the queries, keys, and values as follows:
\begin{equation}
    Q_i = f_i W_i^Q, \quad K_i = f_i W_i^K, \quad V_i = f_i W_i^V
\end{equation}

where $W_i^Q \in \mathbb{R}^{T \times d_k}$, $W_i^K \in \mathbb{R}^{T \times d_k}$, and $W_i^V \in \mathbb{R}^{T \times d_k}$ are projection matrices, with $d_k$ representing the projection dimension, typically defined as $d_k = \frac{D}{M}$, where $M$ denotes the number of attention heads. The Signed Mask-Attention matrix can be expressed as:
\begin{equation}
    \mathbf{A} = \sigma_{t}(\|Q+K\|_1 W_{\emptyset}^T) \odot \sigma_{s}(QK^T)
\end{equation}

where $\sigma_{t}$ denotes the tanh activation function, which outputs values in the range $[-1, 1]$, and $\sigma_{s}$ denotes the sigmoid activation function, which outputs values in the range $[0, 1]$. Additionally, $W_{\emptyset}$ is a learnable matrix with the same dimensions as $Q$. These activation functions introduce non-linearity into the learning process while also constraining the matrix values to prevent them from becoming excessively large. Our newly designed attention matrix leverages the information in $Q$ and $K$ more effectively, adjusting the magnitude of the original attention mechanism. This approach emphasizes important areas while downplaying unimportant ones, enhancing the model's focus on critical regions.
\begin{equation}
    \text{SMA}(Q_i, K_i, V_i) = \text{Softmax}\left(\mathbf{A_i} / d_k \right)\times V_i
\end{equation}

Subsequently, this result is added to the original features, followed by the application of a feedforward network. The output is then added again to obtain the final vector. Therefore, the SMA branch can be computed as:
\begin{equation}
f_{sma} = \text{FFN}(\text{LN}(\text{SMA}(DS(f)) + DS(f)))
\end{equation}
where LN is the LayerNorm, DS is the down-sampling layer, FFN is the FeedForward Network. 

Additionally, we designed a convolutional network-based branch to more effectively extract the high-frequency information from the input signal. This is accomplished through the use of convolution and max-pooling operations. Specifically, we have:
\begin{equation}
    f_{pool} = Conv(\sigma_{s}(\text{Maxpool}(f)) \odot \text{DS}(f)),
\end{equation}

Finally, we employ a mixed module that combines channel-wise concatenation with ContraNorm~\cite{guo2023contranorm} to aggregate the features from the two distinct branches. The ContraNorm operation has been demonstrated to effectively disentangle representations in the embedding space, thereby enhancing generalization performance.

\begin{equation}
\begin{aligned}
    f_c &= Conv_{1x1}(Cat[f_{sma}, f_{pool}])\\
    f_{TSSE} &= f_{c} - {\tau} \cdot \text{softmax}(f_{c} \times f_{c}^\top) f_{c}
    \end{aligned}
\end{equation}

where $Conv_{1x1}$ denotes a convolutional layer with a kernel size of 1 and a stride of 1, the channel is set as D. The channel-wise concatenation module is used to fuse $f_{ssa}$ and $f_{pool}$,  $\tau$ is the parameter.

Our TSSE design allows our network to leverage the unique strengths of each frequency band, thus enhancing the precision and robustness of our temporal activity detection system. By fusing the separately learned high and low-frequency features, we create a more comprehensive representation of the signal. This fused representation effectively combines the detailed temporal boundaries derived from the high-frequency features with the contextual insights from the low-frequency features. As a result, our model excels in accurately identifying both the timing and nature of activities, even amidst complex and noisy environments, as evidence by the numerical results in the experimental section.


\subsubsection{Local Sensitive Response Encoder}

We further design another encoder to capture the fluctuation information to enhance the localization. As depicted in Fig. \ref{fig:framework}, the LSRE employs a channel-wise window to slide on temporal axis to extract regional information. The scale of window is determined by the LSRE order $l$. To capture the regional fluctuations, we compute the maximum and minimum values within the window to obtain the aggregated feature. This process can be formalized as:

\begin{equation}
\small
    f_{w} =Concat_t \{\max(f_{in}[t:t+2^l])\ - \min(f_{in}[t:t+2^l])\}_{t=0}^{T-w}
\end{equation}

where $f_{\text{w}} $ represents the aggregated feature, and the $t$ denotes the window's position, $Concat_t$ means to concatenate the features in the temporal dimension. The sliding window effectively captures the local fluctuation by computing the difference between the maximum and minimum values within each window, thereby enhancing the regional saliency of the input signal. Most importantly, this operation is achieved in a learning-free manner. In this way, it reduces complexity and computational overhead, enabling faster processing and lower memory usage. It’s also less prone to overfitting, making it an efficient and generalizable method for capturing essential signal characteristics.
For the aggregated features, a MLP layer is utilized to transform the raw regional difference into a more robust and discriminative feature space
\begin{equation}
    f_{\text{LSRE}} = \text{MLP}(f_{\text{w}})
\end{equation}

where the output feature $f_{\text{LSRE}} $ encapsulates crucial regional saliency information, serving as a robust and informative input for subsequent processing.

\subsubsection{Cross-attention Pyramid Fusion}
By obtaining the outputs from each encoder, we construct the feature pyramids. One pyramid contains the multi-scale output of TSSE, denoted as $Set_T = \{f^1_{TSSE}, f^2_{TSSE}, \ldots, f^L_{TSSE}\}$, and the other is the LSRE pyramid, denoted as $Set_L = \{f^1_{LSRE}, f^2_{LSRE}, \ldots, f^L_{LSRE}\}$. The scale of these pyramids is determined by a stride of 2. To efficiently and comprehensively fuse these two pyramid sets, we designed the Cross-Attention Pyramid Fusion, which aims to align and integrate the feature information from both pyramids. Specifically, for the features at the $l^{th}$ level, this process can be formulated as:
\begin{equation}
\begin{cases}
f^l_{c1} \leftarrow \text{CrossAttention}(f^l_{LSRE}, f^l_{TSSE}), \\
f^l_{c2} \leftarrow \text{CrossAttention}(f^l_{TSSE}, f^l_{LSRE})
\end{cases}
\end{equation}

where $fl_{LSRE}$ represents the feature at the $l^{th}$ level of the LSRE pyramid, and $f^l_{TSSE}$ represents the corresponding feature at the  $l^{th}$ level of the TSSE pyramid. The CrossAttention operation between these features enables the model to capture and exchange complementary information from both pyramids. This bi-directional attention mechanism ensures that the fused features $f^l_{c1}$ and $f^l_{c2}$ incorporate both local and global context, enhancing the overall representation capability of the model.

After that, we have post-processed the two features and add additional $f^l_{TSSE}$ through layernorm for facilitating training
\begin{equation}
    f^l_{det} = {LN(FFN(f^l_{c1}+f^l_{c2}) + LN(f^l_{TSSE})}
\end{equation}

where $f^l_{det}$ represents the final fused feature used for detection at the $l^{th}$ level of the pyramid. 


\subsection{Prediction Head}

We build the prediction head to process the pyramid features across multi levels. It can be divided into two symmetric classification branch and localization branch which are both realized by 3 CGR layers with a single convolution layer. The difference between the two branches is the classification branch conv projects the feature $f^l_{det} \in \mathbb{R}^{T^l \times {D}}$ into the class score predictions $f^l_{cls} \in \mathbb{R}^{T^l \times {cls}}$ and localization branch conv projects the feature $f^l_{reg} \in \mathbb{R}^{T^l \times {2}}$ into the boundary locations, respectively at different time stamps. 

For an instant $t^l$ in the $l_{th}$ level, the prediction head estimates the boundary distance $\hat{d}^l_{st}$ and $\hat{d}^l_{et}$, the class scores $\hat{c}^l_t$ of all categories with the background is also obtained. Then the candidate activity segments $\hat{o}^l_t=(\hat{a}^l_t,  \hat{s}^l_t, \hat{e}^l_t)$ can be decoded by

\begin{equation}
\hat{a}^l_t = \arg \max (\hat{c}^l_t), \quad \hat{s}_t^l = t^l - \hat{d}^l_{st}, \quad \hat{e}_t^l = t^l + \hat{d}^l_{et}
\end{equation}

\subsection{Training and Inference}
\subsubsection{Training} In the training stage, the network outputs predicted candidates $\hat{o}$, we optimize the model by aligning the predicted results with the ground truth annotations. The objective function of the proposed DPWiT optimization follows the design in \cite{zhang2022actionformer}, which has two sub-functions, the first sub-function $\mathcal{L}_{\text{Cls}}$ is a focal loss\cite{lin2017focal} for classification, the second sub-function $\mathcal{L}_{\text{Loc}}$ is a DIoU loss\cite{zheng2020distance} for distance regression. The objective function is defined as

\begin{equation}
    \mathcal{L} = \sum_{t} (\mathcal{L}_{\text{Cls}} + \lambda\mathcal{L}_{\text{Loc}}) / T_+
\end{equation}

where $T_+$ is the number of positive predictions and $\lambda$ is an hyper-parameter to modulate the ration of $\mathcal{L}_{\text{Cls}}$ and $\mathcal{L}_{\text{Loc}}$.

\subsubsection{Inference} At inference stage, the predicted candidates $\hat{o}$ with classification scores higher than threshold $\beta$ and their corresponding instances are kept. We then assemble all predictions and process them with Soft-NMS\cite{bodla2017soft} to duplicate overlapped instances.

\section{Experiments}
\label{sec:exp}
\subsection{Dataset}
We collected CSI samples to create the dataset used in our experiments. The CSI data was gathered in an empty office room measuring $7m \times 12m \times 2.5m$. Our test bed consists of two laptops equipped with commercial Intel 5300 NICs, functioning as the transmitter (TX) and receiver (RX), respectively. Both the TX and RX each have one antenna, with thirty sub-channels available. Three student volunteers participated in the experiment, where they were asked to randomly perform a set of predefined daily activities between the TX and RX. The key statistics of our dataset are summarized in Table \ref{tab:dataset_info}. The device's sampling frequency is 100Hz. The signal recordings cover seven daily activities—walking, running, jumping, waving, falling, sitting, and standing—and include 553 untrimmed samples with a total of 2,114 activity instances. Each instance is meticulously annotated with its start time, end time, and activity category. The whole dataset is split with a 7:3 ratio as the training and testing subsets.
\begin{table}[h]
    \centering
    \scriptsize
    {
    \begin{tabular}{c|c|c|c|c|c|c|c}
        \hline
        Category & Walk & Run & Jump & Wave & Fall & Sit & Stand \\
        \hline
        Num. & 394 & 361 & 347 & 335 & 332 & 225 & 120 \\
        Avg.(s) & 16 & 17 & 13 & 18 & 13 & 13 & 11 \\
        Max.(s) & 30 & 25 & 20 & 30 & 25 & 40 & 20 \\
        Min.(s) & 10 & 5 & 5 & 10 & 5 & 5 & 5 \\
        \hline
    \end{tabular}
    }
    \caption{Details of the dataset}
    \vspace{-0.5cm}
    \label{tab:dataset_info}
\end{table}

\begin{table*}[!t]
\begin{center}
\label{tab:performance}
\resizebox{\textwidth}{!}{%
\begin{tabular}{l|ccccc|>{\columncolor{lightgray}}c|ccc}
    \toprule
    Model & $\text{mAP}_{0.3}$ & $\text{mAP}_{0.4}$ & $\text{mAP}_{0.5}$ & $\text{mAP}_{0.6}$ & $\text{mAP}_{0.7}$ & $\text{mAP}_{avg}$ & GFlops & Time/ms \\
    \midrule
    Baseline-ResNet1d\cite{he2016deep} & 19.7 & 19.2 & 16.7 & 10.9 & 7.5 & 14.8 & 0.29 & \textbf{7.8} \\
    Baseline-THAT\cite{li2021two} & 21.2 & 20.6 & 16.2 & 10.6 & 6.7 & 15.1 & 1.64 & 8.2 \\
    \midrule
    AFSD\cite{lin2021learning} & 46.6 & 45.4 & 42.4 & 37.9 & 25.4 & 39.5  & 91.0 & 71.8 \\
    BREM\cite{brem} & 48.8 & 46.4 & 43.5 & 38.7 & 29.7 & 41.4  & 205.1 & 74.7 \\
    ActionFormer\cite{zhang2022actionformer} & 60.8 & 58.6 & 56.6 & 50.5 & 35.8 & 52.5  & 148.3& 98.5 \\
    TADTR\cite{liu2022end} & 63.3 & 59.9 & 57.4 & 51.8 & 38.2 & 54.1  & 185.4 & 75.7 \\
    Tridet\cite{shi2023tridet} & 62.4 & 60.8 & 58.5 & 53.8 & 39.5 & 55.0  & 239.5 & 117.4 \\
    TemporalMaxer\cite{tang2023temporalmaxer} & 64.5 & 62.0 & 60.0 & 54.9 & 40.5 & 56.4  & 189.6 & 76.1 \\DyFADet\cite{yang2024dyfadet} & 66.8 & 64.2 & 62.4 & 56.4 & 40.1 & 58.0  & 304.0 & 106.0 \\
    Ours & \textbf{85.5} & \textbf{83.0} & \textbf{77.3} & \textbf{72.1} & \textbf{54.5} & \textbf{74.5}  & \textbf{44.1} & 61.8 \\
    \bottomrule
\end{tabular}}
\caption{Comparison of the model mAPs, GFLOPs and inference time(ms) of different methods.}
\vspace{-0.1cm}
\label{tab:main}
\end{center}
\end{table*}

\subsection{Experimental Setups}
\subsubsection{Metrics and Baselines} 


We evaluate our model's performance using Mean Average Precision (mAP) at several temporal Intersection over Union (tIoU) thresholds. tIoU is defined as the ratio of the intersection to the union of two temporal windows, determining localization accuracy. If tIoU exceeds a threshold, the window is validated for correct action classification. mAP is then obtained by averaging the Average Precision (AP) across all categories. Following Thumos14~\cite{idrees2017thumos}, tIoU thresholds range from 0.3 to 0.7 in steps of 0.1.

The baseline comparison methods are primarily adapted from the vision community and include the following: \textbf{AFSD} \cite{lin2021learning}, which learns salient boundary features without anchors; \textbf{BREM} \cite{brem}, which predicts multi-scale boundary quality to improve proposal scores; \textbf{ActionFormer} \cite{zhang2022actionformer}, which combines local self-attention with multiscale features for long-range temporal context; \textbf{TADTR} \cite{liu2022end}, which uses temporal deformable attention to focus on key snippets; \textbf{Tridet} \cite{shi2023tridet}, which models action boundaries with a relative probability distribution; \textbf{TemporalMaxer} \cite{tang2023temporalmaxer}, which emphasizes feature extraction while minimizing long-term context modeling; and \textbf{DyFADet} \cite{yang2024dyfadet}, which introduces Dynamic Feature Aggregation to adapt kernel weights and receptive fields over time. To ensure fair comparison, we modified their pipelines to fit signal data and replaced their backbones with the same encoder used in our network. Additionally, we implemented two sliding-window baseline methods: one based on a Convolutional Network and the other on a Transformer. The input is pre-segmented according to ground truth annotations, with a window size matching the maximum ground truth boundary. These networks are trained to classify activity categories and background, and during inference, a sliding window approach is used to identify activities within each segment.

\subsubsection{Implementation Details} The model is implemented in PyTorch, using Adam as the optimizer with an initial learning rate of 4e-5 and a weight decay coefficient of 1e-3. Training was conducted on a workstation equipped with an Intel(R) Xeon(R) Gold 5218R CPU @ 2.10GHz and two Nvidia 3090 GPUs, with a batch size of 2. Training our dataset required 40 epochs, taking approximately 4 hours to complete. During the inference stage, the model outputs were processed by Soft-NMS, with a sigma value of 0.95 and a confidence threshold of 0.01. For both training and inference, single signal samples were divided into clips, each with a length of 4096 time stamps (approximately 41 seconds, covering over 2 activities), with a stride of 0.5. We utilized 8 TSSE and LSRE backbones as feature encoders, and the output features from the last 4 layers were used for detection. Regarding the hyperparameters, the coefficient $\lambda$ in the objective function was set to 10, the scale $\tau$ in ContraNorm was set to 0.1, and the confidence threshold in focal loss $\beta$ was set to 0.9.

\begin{table}[t]
\centering
\resizebox{\linewidth}{!}
{
\begin{tabular}{lcccc}
\toprule
Method & $\text{mAP}_{0.3}$ & $\text{mAP}_{0.5}$ & $\text{mAP}_{0.7}$ & $\text{mAP}_{avg}$ \\
\midrule
\multicolumn{5}{c}{\textit{\textbf{Decomposed Frequency-aware Learning }}} \\
w/o Transformer-based branch & 77.8 & 71.9 & 49.9 & 68.6   \\
w/o Conv-Pool based branch & 78.9 & 70.7 & 46.2 & 66.6  \\
\midrule
\multicolumn{5}{c}{\textit{\textbf{Dual Encoder}}} \\
LSRE & 33.1 & 26.6 & 12.3 & 24.7  \\
TSSE & 79.0 & 70.5 & 44.6 & 66.2  \\
\midrule
\multicolumn{5}{c}{\textit{\textbf{Design in LSRE}}} \\
min & 80.0 & 73.9 & 46.4 & 68.8  \\
mean & 83.6 & 74.1 & 47.9 & 70.3  \\
max & 82.1 & 75.1 & 50.5 & 71.3  \\
max-min & \textbf{85.5} & \textbf{77.3} & \textbf{54.5} & \textbf{74.5} \\
\midrule
\multicolumn{5}{c}{\textit{\textbf{Pyramid Fusion}}} \\
Pyramid-wise Add & 80.8 & 72.3 & 48.6 & 69.0 \\ \midrule
\multicolumn{5}{c}{\textit{\textbf{Signed Mask-Attention}}} \\
Self-Attention & 70.7 & 64.0 & 35.1 & 59.4  \\
\midrule
\multicolumn{5}{c}{\textit{\textbf{Proposed Method}}} \\
\rowcolor{lightgray}\textit{\textbf{Ours}} & \textbf{85.5} & \textbf{77.3} & \textbf{54.5} & \textbf{74.5} \\
\bottomrule
\end{tabular}}
\caption{Ablation studies from various aspects.}
\vspace{-0.5cm}
\label{tab:ablation}
\end{table}

\subsection{Main Results}
We report our main results in Table~\ref{tab:main}, which demonstrate that our model outperforms all baselines and achieves state-of-the-art performance on the dataset. Notably, DyFADet achieves close to $60\% $ accuracy, while TemporalMaxer, using MaxPooling for dimensionality reduction, loses semantic information, resulting in a lower mAP of $56.4\%$. Single-stage models like AFSD and BREM achieve $39.5\%$ and $41.4\%$ mAP, respectively, while TadTR, adapted from DETR, scores $54.1\%$. These results highlight the critical role of model design in temporal activity detection using WiFi-based data. Cross-person evaluation is provided in the supplementary material.

\begin{figure}[t]
    \centering
    \includegraphics[width=1\linewidth]{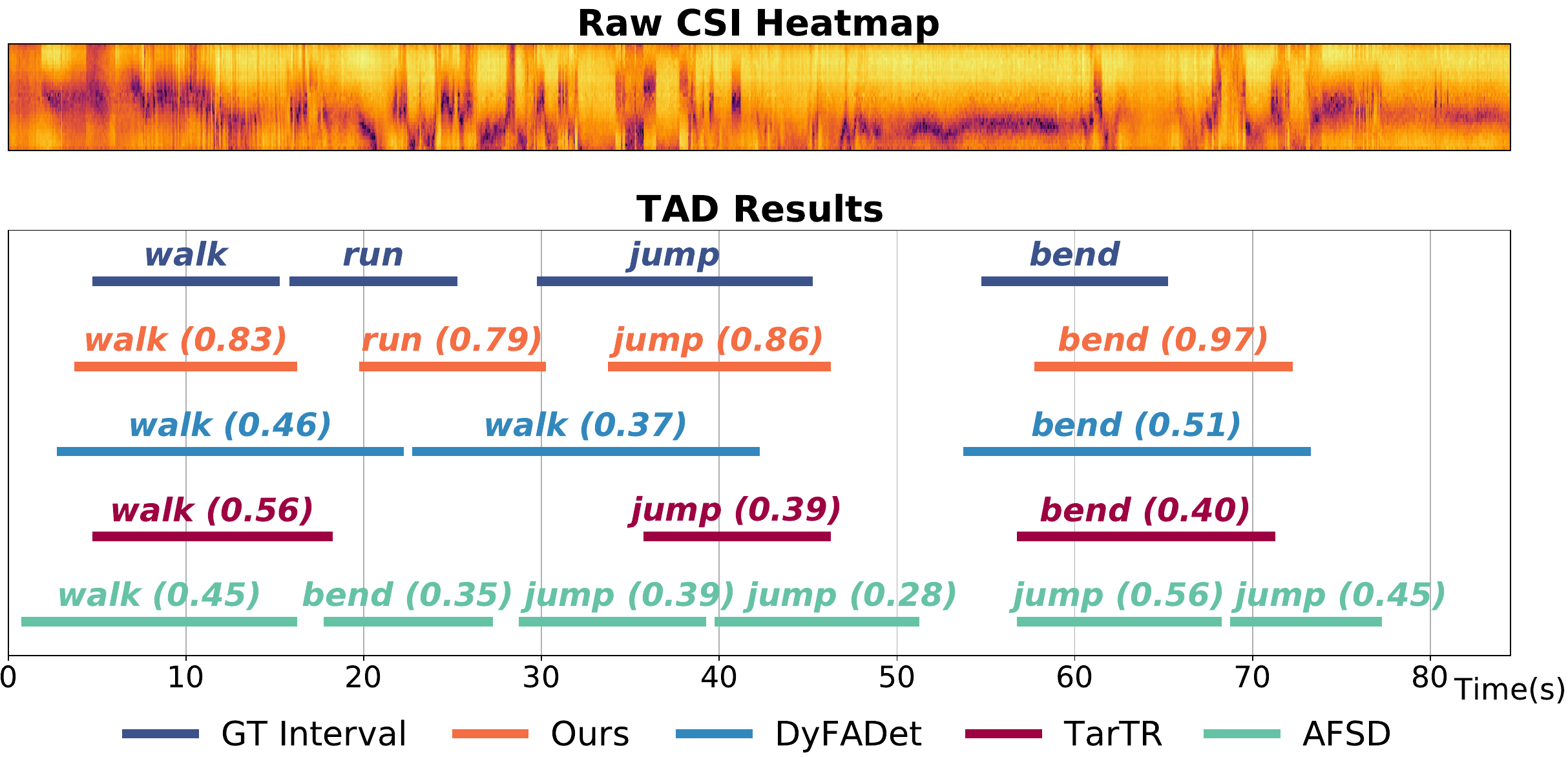}
    \caption{ Prediction Visualizations form different methods.}
    \label{fig:visual}
\end{figure}

\subsection{Ablation Study}

To further verify the efficacy of our contributions, we conduct extensive ablation studies on Dataset for our method in Table \ref{tab:ablation}.

First, we evaluated the benefits of the \textbf{decomposed frequency-aware learning} mechanism. Removing either the Transformer or Convolution-pooling branch significantly reduced performance, confirming the value of high and low-frequency decomposition. Second, we assessed the impact of \textbf{Dual Encoders}. While the system with only LSRE struggled, combining it with TSSE showed substantial improvements. Third, we examined the \textbf{Design in LSRE} and found that capturing both maximum and minimum values empirically delivered the best results. Fourth, we analyzed the \textbf{pyramid fusion} mechanism. Conventional methods, like pairwise addition, performed poorly, underscoring the need to model feature interactions. Finally, replacing the proposed \textbf{Signed Mask-Attention} with conventional self-attention led to significantly worse results, demonstrating its effectiveness in focusing on informative input areas.

\begin{table}[b]
\centering
\resizebox{\linewidth}{!}{%
\begin{tabular}{c|p{9cm}}
\toprule
\textbf{Error Type} & \textbf{Definition} \\ 
\midrule
Background & Correctly labeled predictions with an IoU of less than 0.1\\
\midrule
Localization  & Correctly labeled predictions with an IoU between 0.1 and 0.5\\
\midrule
Wrong Label  & Predictions with more than 0.5 IoU but with incorrect class labels\\
\midrule
Confusion & Incorrectly labeled predictions with an IoU between 0.1 and 0.5\\
\midrule
Double Detection & Correctly labeled predictions with an IoU over 0.5, but the GT is already matched with a higher-scoring wrong prediction\\
\bottomrule
\end{tabular}
}
\caption{Definition of error metrics in the False Positive Analysis.}
\label{tab:err}
\end{table}


\begin{figure}[t]
    \centering
    \includegraphics[width=1\linewidth]{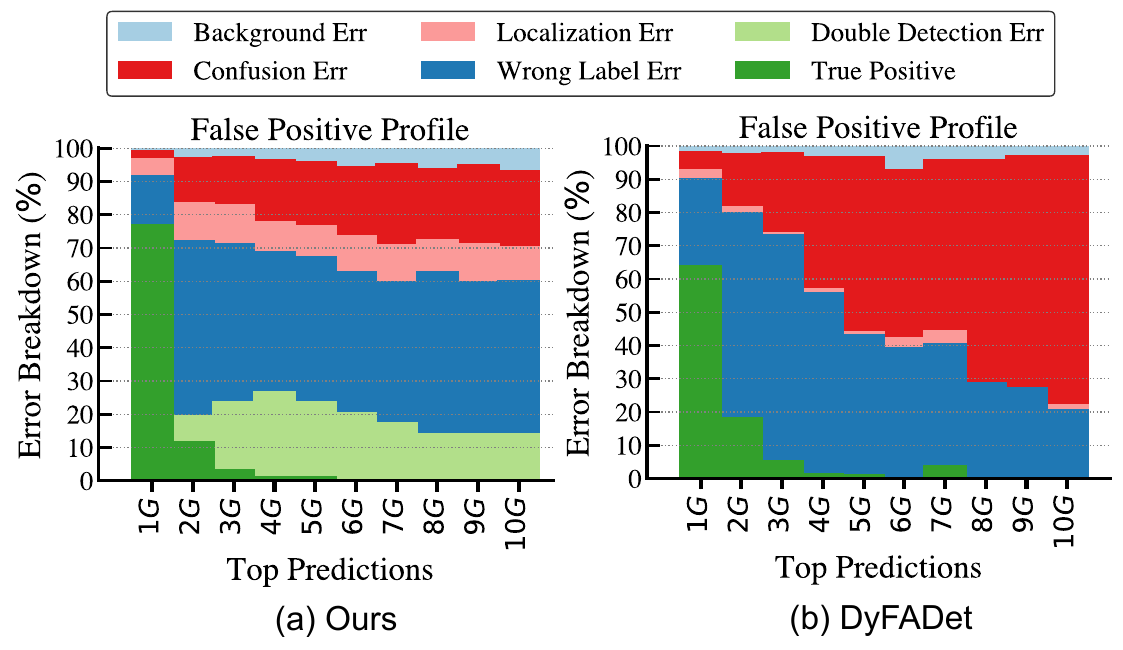}
    \caption{False Positive Profiling on wireless dataset, we use DPWiT as the model and compare our net with DyFADet with their best mAP results.}
    \vspace{-0.5cm}
    \label{fig:FP}
\end{figure}

\subsection{Analyze}
\subsubsection{Visual Analyze}
We present a qualitative visualization of our dataset in Fig. \ref{fig:visual}. The figure shows the ground truth activities in the CSI data alongside the best-predicted proposals from four different models.  As observed, our method accurately identifies candidate actions and provides reliable temporal locations. In contrast, DyFADet correctly identifies the location but misclassifies the action, TADTR misses the location, and AFSD produces scattered predictions that fail to form a coherent result.

\subsubsection{False Positive Analyze}
To further diagnose the errors predicted by our proposed method, we follow the process outlined in~\cite{alwassel_2018_detad} to conduct a False Positive (FP) Analysis. Specifically, this framework involves five error metrics, which are defined in Table~\ref{tab:err}. The results are illustrated in Fig. \ref{fig:FP}. This analysis allows us to evaluate the error profile of the top-10$G$ predictions, where $G$ represents the number of ground truth instances. We select the top predictions in a per-class manner, meaning we choose the top-10$G_j$ predictions from class $j$, where $G_j$ is the number of instances in class $j$. Additionally, to observe the trend of each error type, we divide the top-10$G$ predictions into ten equal splits and examine the breakdown of the five FP error types in each split. Compared to the leading competing method, DyFADet, our method generates more informative predictions that improve positive localization and reduces confusion errors, thereby minimizing false positive judgments.

\section{Conclusion}
\label{sec:conc}

This work tackles the challenging problem of wireless temporal activity detection. We propose a Dual Pyramid Network that integrates high- and low-frequency features via a Temporal Signal Semantic Encoder and refines them using a Local Sensitive Response Encoder and cross-attention pyramid fusion. To support this task, we introduce a dataset with 2,114 activity segments from 553 WiFi CSI samples. Extensive experiments show our method significantly outperforms existing baselines, advancing temporal activity detection.

\section{Acknowledgments}
This work was supported by the Key Program for International Cooperation of Ministry of Science and Technology of China (No.2024YFE0100700) and the National Natural Science Foundation of China (NSFC) under Grant 62020106011. The work was  also supported in part by National Natural Science Foundation of China (NSFC) under Grant No. 62171302, the 111 Project under Grant No. B21044 and Sichuan Science and Technology Program under Grant No. 2023NSFSC1965.

\bibliography{aaai25}
\end{document}


\maketitle

\section{Introduction}
In this supplementary material, we add and discuss the additional frequency-aware evaluation and the cross-person evaluation to clarify our method.

\section{Frequency-aware Evaluation}
To better support our claim on the different sensitivity of Transformer and CNN from the frequency angle, we built one CNN and one Transformer with comparable parameters as the feature extractor. We set 1/4 of the frequency spectrum as the low- and high-frequency boundary and tested each model in low- and high-frequency bands, measuring their sensitivity through reductions in MAP values. Our results are displayed below:

\begin{table}[htbp]
\footnotesize
\centering
\begin{tabular}{c|c|c}
\toprule

\textbf{Frequency Band} & \textbf{MAP Reduction (TF)} & \textbf{MAP Reduction (CNN)} \\
\midrule
(0, 1)   & 0\%      & 0\%       \\
(0, 1/4) & 1.82\%   & 4\%       \\
(0, 1/8) & 7.47\%   & 13.49\%   \\
(0, 1/9) & 9.17\%   & 16.17\%   \\
(0, 1/16) & 29.2\%  & 31.1\%    \\
\bottomrule
\end{tabular}
\caption{Frequency Band Analysis: AMP and MAP Reduction Results}
\label{tab:freq_reduction}
\end{table}


\begin{table}[htbp]
\footnotesize
\centering
\begin{tabular}{c|c|c}
\toprule

\textbf{Frequency Band} & \textbf{MAP Reduction (TF)} & \textbf{MAP Reduction (CNN)} \\
\midrule
(0, 1)      & 0\%       & 0\%        \\
(1/4, 1)    & 83.35\%   & 68.14\%    \\
(1/2, 1)    & 82.99\%   & 70.09\%    \\
(3/4, 1)    & 83.02\%   & 71.10\%    \\
\bottomrule
\end{tabular}
\caption{Frequency Band Analysis: AMP and MAP Reduction Results for Higher Bands}
\label{tab:freq_reduction_high}
\end{table}

The results indicate that the Transformer is more sensitive to low-frequency information, while the CNN is more responsive to high-frequency variations.

\section{Cross-Person Evaluation}

In the main paper, we did not include cross-person evaluation. Hence, we conduct complementary evaluation on unseen subjects here. We conducted leave-one-group-out validation, following the protocol of using 70\% for training and 30\% for testing. Ten volunteers are divided into a fixed 3:3:4 combination with non-overlapping IDs. Two groups ("3+3") are used for training, and the remaining group ("4") for testing, simulating unseen samples. We compare our model with Dyfadet, ActionFormer, and THAT, repeating the procedure three times. The summarized results are shown in the table below. Results clear show the consistent advantage of the proposed method against the baseline methods.
\begin{table}[!thb]
\footnotesize
\centering
\begin{tabular}{c|c|c|c|c}
\toprule
\textbf{Metric} & \textbf{THAT} & \textbf{Dyfadet} & \textbf{Actionformer} & \textbf{Ours} \\
\midrule
\textbf{mAP0.3} & 18$\pm$0.81   & 65.91$\pm$7.33 & 59.91$\pm$6.1  & 75.05$\pm$3.56 \\
\textbf{mAP0.4} & 17.36$\pm$0.84  & 63.79$\pm$8.44  & 57.41$\pm$6.8   & 72.01$\pm$5.83 \\
\textbf{mAP0.5} & 14.28$\pm$0.96  & 60.1$\pm$9.34   & 54.49$\pm$7.95  & 68.15$\pm$6.09 \\
\textbf{mAP0.6} & 8.45$\pm$1.56   & 50.72$\pm$6.42  & 45.9$\pm$6.61   & 61.16$\pm$5.88 \\
\textbf{mAP0.7} & 5.04$\pm$0.42   & 33.01$\pm$3.34  & 31.7$\pm$4.1    & 44.36$\pm$4.26 \\
\textbf{Avg.}   & 12.63$\pm$0.77  & 54.71$\pm$6.89  & 49.88$\pm$6.29  & 64.15$\pm$5.1  \\
\bottomrule
\end{tabular}
\caption{Performance Comparison Across Different Models}
\label{tab:map_comparison}
\end{table}